# Fine-Grained Decision-Theoretic Search Control

Stuart Russell
Computer Science Division
University of California
Berkeley, CA 94720

## Abstract

Decision-theoretic control of search has previously used as its basic unit of computation the generation and evaluation of a complete set of successors. Although this simplifies analysis, it results in some lost opportunities for pruning and satisficing. This paper therefore extends the analysis of the value of computation to cover individual successor evaluations. The analytic techniques used may prove useful for control of reasoning in more general settings. A formula is developed for the expected value of a node, $k$ of whose $n$ successors have been evaluated. This formula is used to estimate the value of expanding further successors, using a general formula for the value of a computation in game-playing developed in earlier work. We exhibit an improved version of the MGSS* algorithm, giving empirical results for the game of Othello.

## 1 Introduction

In earlier work with Eric Wefald [1988, 1989, in press], the author developed an approach to controlling computation based on maximizing the expected value of computation. The method involves dividing the base-level decision-making process into atomic steps, such that the step with the highest expected value is taken at each juncture until the value of further computation is negative. The resulting algorithms for single-agent search and game-playing have exhibited good performance. However, several restrictions were imposed to simplify the analysis. One such restriction identified the computation steps in game-playing with the complete one-ply expansion of a leaf node, rather than allowing the program to control the generation of individual successors. This simplification has several advantages, including the fact that the nodes in the tree have well-defined values at all times when min or min-max backup is used.

On the other hand, some opportunities for pruning are lost, in particular those opportunities taken by alpha-beta search to stop generating successors as soon as the node is found to be valueless. Satisficing effects are also lost. These come into play when a node has a large number of successors; it is often necessary to examine only a small number of them in order to get a good estimate of the value of the node [Pearl, 1988].

In this paper, I attempt to rectify the situation by extending the analysis of the value of computation to the case of single successor generation and evaluation. To do this, the following steps are followed:

1. Derive a formula for the expected value of a node when only a subset of its successors have been evaluated.

2. Use this formula to estimate the value of expanding further successors, using the general formula for the value of a computation in game-playing [Russell and Wefald, 1989].

3. Derive pruning conditions, under which a node's expansion must have zero expected benefit.

4. Using the formula from step 2, and the pruning conditions from step 3, implement the algorithm and demonstrate its performance.

The sections of the paper parallel these steps, more or less. While the formulæ developed look quite formidable, the basic ideas are straightforward, and a number of qualitative insights are obtained.

## 2 The value of a partially expanded node

Consider a min-node $j$, $k$ of whose $n$ successors have been generated and evaluated, yielding values $u_1 \ldots u_k$. The expected value of the node $j$ is clearly less than or equal to $min_k(j) = min(u_1 \ldots u_k)$. Alpha-beta search uses this inequality to prune certain nodes before all their successors have been generated. For a calculation of the expected value of generating and evaluating one more successor, the $k+1^{th}$, we need a more precise description of the expected value of the node. To do this, we calculate a probability distribution for the value of the minimum of the $n$ successors, given that $k$ of the $n$ have already been evaluated.

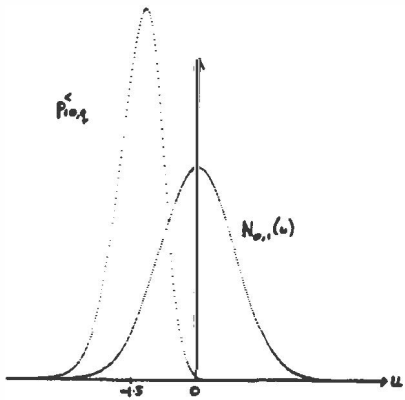

Figure 1: $p_{10,q}^<$ for $q$ = standard normal curve $N_{0,1}$

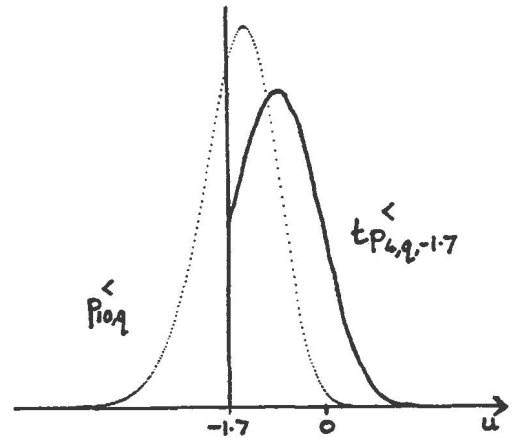

Figure 2: $tp_{4,q,-1.7}^<$, the density function after two successors

The desired distribution is essentially a truncated version of the general distribution for the minimum of some number of random variables. The latter distribution, which we shall call the $p^<$-distribution (pronounced 'p-min'), depends on the distribution from which the random variables are drawn. Let us assume that the successor values are drawn at random from a distribution $q$, which is specific to the node being expanded. Define the probability distribution $p_{n,q}^<$ as the density function of the minimum of $n$ random variables $U_1 \ldots U_n$ drawn from a distribution $q$. The relation between $q$ and $p_{n,q}^<$ is most easily seen by examining the cumulative distribution function for $p^<$. The derivation is standard in the area of order statistics.[1]

$$\begin{aligned} P_{n,q}^<(x) &= \int_{-\infty}^{x} p_{n,q}^<(y) dy \\ &= prob(min(U_1,\ldots,U_n) \leq x) \\ &= 1 - prob(min(U_1,\ldots,U_n) > x) \\ &= 1 - prob((U_1 > x) \wedge \ldots \wedge (U_n > x)) \\ &= 1 - (1 - Q(x))^n \end{aligned} \quad (1)$$

where $Q$ is the cumulative distribution function corresponding to $q$. Hence

$$p_{n,q}^<(x) = \frac{d}{dx}[P_{n,q}^<(x)] = nq(x)(1 - Q(x))^{n-1} \quad (2)$$

Figure 1 shows the distribution $p_{n,q}^<(x)$ for $n = 10$ associated with the normal distribution $q(x) = N_{0,1}(x)$.

Now we can easily compute the distribution of the minimum of the $n$ successors, given that $k$ of the $n$ have already been evaluated. If $min_k$ is the minimum value of the first $k$ successors, then the minimum of all $n$ successors is clearly less than or equal to $min_k$. For $x < min_k$, the probability that $min_n = min(u_1,\ldots,u_n)$ lies between $x$ and $x+dx$ is just $p_{n-k,q}^<(x)dx$. We will use the notation $tp_{l,q,m}^<(x)$ ("truncated p-min") for the probability distribution of $min_n$ when there are $l$ unevaluated successors and the minimum of the evaluated successors is $m$. Then

$$tp_{l,q,m}^<(x) = \begin{cases} p_{l,q}^<(x) & \text{if } x < m \\ (1 - P_{l,q}^<(m))\delta(x) & \text{if } x = m \\ 0 & \text{otherwise} \end{cases} \quad (3)$$

where $\delta(x)$ is the unit delta-function at $x$. A typical instance of $tp^<$ is shown in Figure 2, where it is assumed that six successors have been generated, with a minimum value of -1.7.

---

[1] We assume here that the successor values are independent given the distribution. In fact, we could consider updating the parameters of the distribution as successors are observed. While this extension would present no serious difficulty, the results obtained without it are adequate.

We will use the notation $b_{l,q}^<(m)$ to denote the expectation of the $tp^<$ distribution, as a function of the truncation point $m$. A typical instance of $b^<$ is shown in Figure 3, superimposed on its corresponding $p^<$ curve. As $m \to -\infty$, the $b^<$ curve is asymptotic to $y = m$, while as $m \to \infty$, the curve is asymptotic to $y = c$, where $c$ is the expectation of the $p^<$ curve.

The expected value of the node after $k$ successors have been evaluated is therefore given by $b_{n-k,q}^<(min_k)$. The function $b^<$, and its dual function $b^>$, thus replace the standard min and max used in backing up leaf node values towards the root.

## 3 The value of evaluating further successors

In [Russell and Wefald, 1989], general formulæ are given for the expected value of carrying out a computation that affects the estimated value of a node in a game tree. Here we consider the value of further expanding a min-node $j$ that is in the subtree of some top-level move $\beta_i$ other than the current best move (the three other cases are handled analogously). In this case the expected benefit $\Delta(S_j)$ of the computation $S_j$ that expands some successors of $j$ is given by

$$E(\Delta(S_j)) = \int_\alpha^\infty p_{ij}(x)(x-\alpha)\,dx \qquad (4)$$

where $p_{ij}$ is the density function for the new value of the top-level move $\beta_i$ after the computation $S_j$, and $\alpha$ is the value of the current best move.

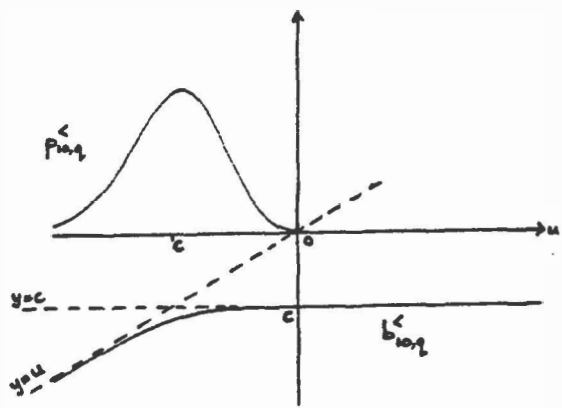

Figure 3: $b_{10,q}^<(m)$ plotted with $p_{10,q}^<$

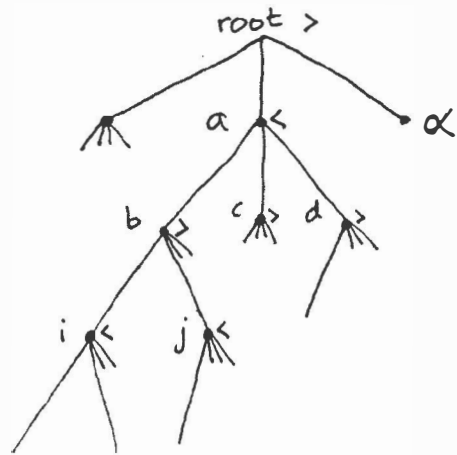

Figure 4: A partial game tree containing partially expanded nodes

### 3.1 How node values are propagated

One of the basic techniques used in this approach is to write the value of the top-level move as a function of the value of the leaf node (and other nodes). Then we can use a standard theorem to rewrite $p_{ij}$ in terms of $p_{jj}$, the density function for the new value of the node $j$.

To derive a formula for the value of the top-level node, I will first go through a particular example, and then ask the reader to take the general formula on faith. Consider figure 3.1, which depicts a partial game search tree in which the root node has been fully expanded, but other nodes are only partially expanded. The computation under consideration is the further expansion of the node labelled $j$; the computation will have value, according to equation 4, only if it causes the value of node $a$ to be raised above $\alpha$, the value of the current best move. In the following I will adopt the notation $b_x^<(m)$ to denote the $b^<$ distribution associated with node $x$ — that is, $b_{l,q}^<$ where $l$ is the number of unexpanded successors of $x$ and $q$ is the distribution from which the values of $x$'s successors are drawn. Where no ambiguity arises, I will use the name of the node to denote its current value also.

First consider the value of node $a$. As explained in the previous section,

$$a = b_a^<(min(b,c,d))$$

Now since $b^<(m)$ is monotonically increasing in $m$, $a$ can only increase its value if $b$ has the lowest value among its siblings and has its value increased (we are assuming that the values of $c$ and $d$ are unaffected by an expansion of $j$). It therefore makes sense to rewrite $min(b,c,d)$ as $min(bound(a),b)$ where $bound(a)$ is the minimum value of the other successors of $a$. Thus

$$a = b_a^<(min(bound(a),b))$$

Going down the tree, we can replace $b$ by the backed-up value of its successors:

$$a = b_a^<(b) = b_a^<(min(bound(a),b_b^>(max(i,j))))$$

Again, we can write $max(i,j) = max(bound(b),j)$, giving us

$$a = b_a^<(min(bound(a),b_b^>(max(bound(b),j))))$$

In general, the $max$ and $min$ will intervene in the composition of $b^<$'s and $b^>$'s at every max and min node respectively on the path from $j$ to the top level. The general formula for a top-level node $n_1$ in terms of a min-node $j$ at depth $2d+1$ is therefore

$$\begin{aligned}n_1 &= b_{n_1}^<(min(bound(n_1),\ldots\\ &\quad b_{n_{2i}}^>(max(bound(n_{2i}),\\ &\quad b_{n_{2i+1}}^<(min(bound(n_{2i+1}),\ldots\\ &\quad b_{n_{2d}}^>(max(bound(n_{2d}),j))..))))..)) \end{aligned} \quad (5)$$

where $n_1\ldots n_{2d}$ are the nodes on the path from the root to $j$. This expression will be denoted by $f(j)$,

so that $f$ is the propagation function from a node to the first level of the tree.[2] Note that in the standard minimax algorithm, the propagation function is identical except that $b^<$ becomes $min$, etc.

### 3.2 The new value of the node being expanded

The computation step $S_j$ will involve evaluating successors $k+1$ through $k+s$, adjusting the estimated value of the node $j$, and propagating the effects of that adjustment. After the expansion, $j$ will have $t = n - k - s$ successors remaining unexamined.

The values $u_{k+1} \ldots u_{k+s}$ of the $s$ successors will be drawn from the distribution $q$ associated with $j$. Let $m_s = min(u_{k+1}, \ldots, u_{k+s})$. Then after drawing, the value of the node $j$ will be given by

$$j = g(m_s) = b^<_{t,q}(min(min_k(j), m_s))$$
$$= \begin{cases} b^<_{t,q}(m_s) & \text{if } m_s < min_k(j) \\ b^<_{t,q}(min_k(j)) & \text{otherwise} \end{cases} \quad (6)$$

(going back to the full notation for $b^<$). Thus $j$, considered as a random variable (the new value of the node after expansion), is a function $g$ of the random variable $m_s$, where $g$ is defined by the above equation.

### 3.3 The value of expanding the node

We are now ready to re-express the density function $p_{ij}$ of the new value of the top-level node (call it $n_1$) from which $j$ is descended. We have

$$n_1 = f(j) = f(g(m_s))$$

where, again, $m_s$ is distributed according to $p^<_{s,q}$. Then following the standard theorem for distributions of functions of random variables, we obtain

$$p_{ij}(x) = p^<_{s,q}(g^{-1}(f^{-1}(x))) \left| \frac{d}{dx} g^{-1}(f^{-1}(x)) \right| \quad (7)$$

Let us consider the sign of the differential expression in this equation. As mentioned above, both $b^<$ and $b^>$ are monotonically non-decreasing, as are $min$ and $max$. Furthermore, monotonicity is preserved under both composition and inversion, hence the differential expression is always non-negative. Therefore, combining equations 4 and 7, we obtain

$$E(\Delta(S_j)) = \int_\alpha^\infty p_{ij}(x)(x - \alpha)\, dx$$
$$= \int_{g^{-1}(f^{-1}(\alpha))}^{g^{-1}(f^{-1}(\infty))} p^<_{s,q}(u)(f(g(u)) - \alpha)\, du \quad (8)$$
$$\text{using } u = g^{-1}(f^{-1}(x))$$

---
[2]Rivest [1988] uses a similar idea to derive his Min/Max Approximation search algorithm.

where the integral range may be restricted to where $g^{-1}$ and $f^{-1}$ are defined. A quick glance back at the definitions of $f$ and $g$ (equations 5 and 6) will be enough to convince the reader that this expression for the value of computation is not that easy to evaluate on the fly; nor does it make immediately clear which nodes will have zero value for expansion. In the next section both deficiencies are more or less remedied.

## 4 Obtaining irrelevance criteria

A node is *irrelevant* to its top-level ancestor if there is no way that a change in the value of the node can change which move is currently regarded as best [Russell and Wefald, 1989]. This criterion is equivalent to the rhs of equation 8 being zero. Basically, this comes about because the $min$'s and $max$'s in the definition of $f$ act as a filter on changes being propagated from the node $j$ whose value is changing, just as in alpha-beta pruning. It is possible to rewrite the expression for $f$ to make this clear, and to yield more explicit irrelevance criteria.[3]

As before, it will be simpler to illustrate the rewriting process on the particular example shown in figure 3.1. First, note that the propagation function is monotonic; therefore only *increases* in node values along the path from $j$ to $a$ are interesting. If any max node on the path (such as $b$) is not the lowest-valued known successor of its parent, then it cannot increase the value of its parent, and is therefore irrelevant. This immediately gives us a pruning test on max nodes.

We will assume, to simplify the exposition, that the above pruning test has been implemented, and we are therefore considering expanding a node $j$ such that all of its max ancestors are lowest known successors. In the case of figure 3.1, we get a simplified expression for $a$, for values of $b$ close to its current value:

$$a = b^<_a(b^>_b(max(bound(b), j)))$$

To rewrite this, we can use the identity

$$b^>_n(max(x,y)) = max(b^>_n(x), b^>_n(y))$$

and other analogous identities for $b^<$ and for $min$. It should be noted that these identities hold for any monotonic function, including arbitrary compositions of $b^<$ and $b^>$. We have

$$a = b^<_a(b^>_b(max(bound(b), j)))$$
$$= b^<_a(max(b^>_b(bound(b)), b^>_b(j)))$$
$$= max(b^<_a(b^>_b(bound(b))), b^<_a(b^>_b(j)))$$

---
[3]Obviously, it is possible to derive these irrelevance criteria without recourse to an expression such as $f(j)$, as is usually done for alpha-beta search. However, as well as ensuring that no opportunities for pruning are missed — as deep cutoffs were missed in alpha-beta — the more formal analysis should be helpful in more complex cases, such as probabilistic games, where informal argument runs out of steam.

Thus we see that $j$ will only affect the top-level node if its new value exceeds $bound(b)$. We can apply the same rewriting process to the general expression for $f(j)$, obtaining

$$n_1 = max( \quad b_{n_1}^<(b_{n_2}^>(bound(n_2))),$$
$$b_{n_1}^<(b_{n_2}^>(b_{n_3}^<(b_{n_4}^>(bound(n_4))))),$$
$$\ldots$$
$$b_{n_1}^<(b_{n_2}^>(b_{n_3}^<(b_{n_4}^>(\ldots b_{n_{2d}}^>(j))))) \quad (9)$$

Thus for its effect to be felt at the top level, the value of $j$ must exceed a series of lower bounds derived from the max nodes on the path to the root. Call the maximum of these bounds $\gamma(j)$. Intuitively, $\gamma(j)$ is the level that the value of $j$ must reach in order to have any effect on its top-level ancestor. Since the value of the $j$ can never exceed the current minimum of its successor values, then if this value is less than $\gamma(j)$ we can deduce that $j$ and all its successors are irrelevant. This gives a second simple pruning test for relevant nodes.

Assuming that the node $j$ is relevant under the criteria above, we can now derive an upper bound on the amount by which the top-level node can change its value. To cut a long story short, the original expression is simplified to

$$n_1 = min( \quad b_{n_1}^<(bound(n_1)),$$
$$b_{n_1}^<(b_{n_2}^>(b_{n_3}^<(bound(n_3))))),$$
$$\ldots,$$
$$b_{n_1}^<(b_{n_2}^>(b_{n_3}^<(b_{n_4}^>(\ldots b_{n_{2d}}^>(j)))))(10)$$

The minimum bound from the min nodes other than $j$ is called the $\delta$ bound [Russell and Wefald, 1989] of the node $j$. Intuitively, $\delta(j)$ is the highest value to which an unbounded increase in $j$ can raise its top-level ancestor. Clearly, if $\delta(j)$ is lower than $\alpha$, then $j$ and all its descendants are irrelevant.

To summarize, a node that is not a descendant of the current best move must pass four tests to be counted as relevant:

1. Its parent must be relevant (unless the parent is the root).
2. Max nodes must be lowest known successors.
3. Min nodes must be able to exceed their $\gamma$ bound.
4. The $\delta$ value of the node must exceed $\alpha$.

If these conditions are met, then we can write

$$f(j) = b_{n_1}^<(b_{n_2}^>(b_{n_3}^<(\ldots b_{n_{2d}}^>(j))))$$

for $f^{-1}(\alpha) \leq j \leq f^{-1}(\delta)$. Thus all the conditionals ($min$'s and $max$'s) are eliminated, and we can write

$$E(\Delta(S_j)) = \int_{g^{-1}(f^{-1}(\alpha))}^{g^{-1}(f^{-1}(\delta))} p_{s,q}^<(u)(f(g(u)) - \alpha) \, du$$

$$+ \int_{g^{-1}(f^{-1}(\delta))}^{g^{-1}(f^{-1}(\infty))} p_{s,q}^<(u)(\delta - \alpha) \, du \quad (11)$$

for the range in which $g^{-1}$ is defined.

The simplified $f$ function is well-behaved, but the $g$ function contains a discontinuity at $u = min_k(j)$, beyond which $g^{-1}$ is undefined. The expression must therefore be evaluated by cases, depending on where the discontinuity falls relative to $\alpha$ and $\delta$. To cut another long story short, here are the resulting expressions:

I. $\underline{\delta, \alpha > f((min_k(j))}$

$$E(\Delta(S_{j_s})) = 0 \quad (12)$$

II. $\underline{\alpha \leq f(g(min_k(j))) \leq \delta}$

$$E(\Delta(S_{j_s})) = f(g(min_k(j)))[1 - P_{s,q}^<(min_k(j))]$$
$$- \alpha[1 - P_{s,q}^<(b_{t,q}^{<^{-1}}(f^{-1}(\alpha)))]$$
$$+ \int_{b_{t,q}^{<^{-1}}(f^{-1}(\alpha))}^{min_k(j)} p_{s,q}^<(u)f(b_{t,q}^<(u)) \, du \quad (13)$$

III. $\underline{\delta, \alpha < f(g(min_k(j))}$

$$E(\Delta(S_{j_s})) = \delta[1 - P_{s,q}^<(b_{t,q}^{<^{-1}}(f^{-1}((\delta))))]$$
$$- \alpha[1 - P_{s,q}^<(b_{t,q}^{<^{-1}}(f^{-1}(\alpha)))]$$
$$+ \int_{b_{t,q}^{<^{-1}}(f^{-1}(\alpha))}^{b_{t,q}^{<^{-1}}(f^{-1}(\delta))} p_{s,q}^<(u)f(b_{t,q}^<(u)) \, du \quad (14)$$

There are three other sets of equations analogous to these, for max nodes relevant to top-level moves other than the current best, and for min and max nodes relevant to the current best move.

## 5 Implementation

The basic structure of the search algorithm is identical to that of the MGSS* algorithm described in [Russell and Wefald, 1989]:

1. Keep taking the computational action with the highest expected net value (benefit minus cost), until none has positive value.
2. Take the move that has the highest expected value after the computations in 1.

The algorithm was implemented by replacing the function that generated all the successors of a node by one that generates just one more successor. The function for computing the value of such an expansion is a direct implementation of the above formulæ, with some approximations: the function returns the highest expected benefit per successor expanded. The algorithm maintains $b$ values for all relevant nodes, and decides relevance as new nodes are generated using the criteria given above. Finally, the backing-up function, formerly $min$ or $max$, is replaced by $b^<$ or $b^>$.

All the distributions and backing-up functions can be calculated exactly in terms of standardized distributions — those derived from $q = N_{0,1}$, the standard normal curve. The standardized distributions are tabulated offline or computed from an exact formula.

The only difficulty that arises is in computing the integral of $p^<(u)f(b^<(u))$. It is possible to provide a tabulated value when $f$ consists of only a single application of $b^<$ or $b^>$, but deeper nestings make the tables unrealistically large. Examination of the $b^<$ and $b^>$ functions, from which $f$ is composed, shows that at the low (or high, respectively) end they act as identity functions; thus the significant filtering comes from the high (or low, respectively) end. The program therefore approximates $f$, for the purposes of the integration, by the $b^<$ or $b^>$ function that exhibits the most critical filtering. We are also experimenting with other methods of approximation.

The source of all the node-specific functions is the mysterious $q$ distribution from which successor values are drawn. The MGSS* algorithm used statistical data, gathered for Othello, of the relative value of the *best* successor. This corresponds to the $p^<$ function. Given the mean and standard deviation of this function, it is possible to calculate the mean and standard deviation of $q$ directly.

## 6 Performance

The algorithm, which I shall call MGSS2, is tested by playing an Othello program using it against a standard alpha-beta search using the same evaluation function, which is a slight improvement over that used by BILL [Lee and Mahajan, 1988]. Our results against alpha-beta search to depths 2 and 6 are summarized in table 1. For each search depth, the time cost function of MGSS* was adjusted to allow the algorithm to play at least as well as alpha-beta. In each tournament, games were played from different randomly-generated starting positions, with the two algorithms alternately playing black. The preliminary results show a significant improvement over the original MGSS* algorithm, and a distinct advantage over alpha-beta (roughly 60:1 at depth 6).

## 7 Conclusions

This paper has taken the application of decision-theoretic metareasoning to game-playing almost to its logical extreme. The performance improvements over standard algorithms are expected to be quite dramatic even on a small-branching-factor game such as Othello. Because of the satisficing nature of the search value calculations, the technique should be much more effective on more complex games such as Go. The formulæ are also applicable, with slight modification, to probabilistic games such as backgammon, where the branching factor can be in the thousands.

| algorithm | wins | nodes |
|---|---|---|
| MGSS2 | 8 | 9,280 |
| $\alpha$-$\beta$ [2] | 8 | 44,866 |
| MGSS2 | 6 | 76,371 |
| $\alpha$-$\beta$ [6] | 2 | 4,640,323 |

Table 1: Summary of results for Othello

The current algorithm generates each successor at random from the set of successors of a node. Clearly, a plausible move generator could be applied to make the selection non-random, with the result that the value of a node would converge much more rapidly to the value obtained by generating all successors. The tools used above can provide a quantitative analysis of the quality of a plausible move generator, and of its effectiveness in reducing search.

It is to be expected that several of the techniques shown below will be useful in many other information value calculations, and in providing pruning criteria for search algorithms using backing-up methods other than min/max. The derivation of the irrelevance criterion, and the subsequent simplification of the search value formula, depended mainly on the monotonicity of the backing-up function. Since the backing-up function is supposed to be an expected-value calculation [Hansson and Mayer, 1989], any reasonable such function should not decrease the value of a parent if one of its successors becomes more valuable.

The work needed to apply the idea of information value to this problem was strenuous at times. Beyond the specific application, it is hoped that some general insights as to the nature of information value have been obtained. But in complex decision systems for real-world problems, one cannot expect closed-form expressions for information value to be easily obtainable or computable. The value of certain types of computation can, however, be learned empirically by examining the actual improvement in decision quality that results. Wefald and Russell [1989] have reported on early efforts in this direction. The efficacy of such learning depends on having a good understanding of what features of the decision situation are relevant — an understanding that can only develop from doing grunt-work like this.